\documentclass[conference]{IEEEtran}
\IEEEoverridecommandlockouts
\usepackage{cite}
\usepackage{amsmath,amssymb,amsfonts}
\usepackage{algorithmic}
\usepackage{graphicx}
\usepackage{textcomp}
\usepackage{xcolor}
\usepackage{booktabs} 
\usepackage{array}    
\usepackage{siunitx}  
\usepackage{multirow}
\usepackage{adjustbox}
\usepackage{makecell}
\usepackage{hyperref}
\usepackage{orcidlink}
\hypersetup{hidelinks}
\hypersetup{
	colorlinks=true,  
	linkcolor=blue,   
	urlcolor=blue     
}
\def\BibTeX{{\rm B\kern-.05em{\sc i\kern-.025em b}\kern-.08em
    T\kern-.1667em\lower.7ex\hbox{E}\kern-.125emX}}
\begin{document}

\title{DualGeo: A Dual-View Framework for Worldwide Image Geo-localization}

\author{
	\IEEEauthorblockN{
		Junchao Cui\textsuperscript{1,2}, 
		Wenqi Shi\textsuperscript{1,2}, 
		Shaoyong Du\textsuperscript{1,2}, 
		Hang He\textsuperscript{1,2}, 
		Xuanzi Ma\textsuperscript{2}, 
		Hao Tang\textsuperscript{1,2}, 
		Xiangyang Luo\textsuperscript{1,2,\textdagger}\thanks{\textdagger~Corresponding author : Xiangyang Luo, luoxy\_ieu@sina.com}}
	\IEEEauthorblockA{
		\textsuperscript{1} \textit{Henan Key Laboratory of Cyberspace Situation Awareness, Zhengzhou, China} \\
		\textsuperscript{2} \textit{Information Engineering University, Zhengzhou, China}
	}
}

\maketitle

\begin{abstract}
Worldwide image geo-localization aims to infer the geographic location of an image captured anywhere on Earth, spanning street, city, regional, national, and continental scales. Existing methods rely on visual features that are sensitive to environmental variations (e.g., lighting, season, and weather) and lack effective post-processing to filter outlier candidates, limiting localization accuracy. To address these limitations, we propose DualGeo, a two-stage framework for worldwide image geo-localization. First, it establishes a geo-representational foundation by fusing image and semantic segmentation features via bidirectional cross-attention. The fused features are then aligned with GPS coordinates through dual-view contrastive learning to build a global retrieval database. Second, it performs geo-cognitive refinement by re-ranking retrieved candidates using geographic clustering. It then feeds them into large multimodal models (LMMs) for final coordinate prediction. Experiments on IM2GPS, IM2GPS3k, and YFCC4k show that DualGeo outperforms state-of-the-art methods, improving street-level ($<$1~km) and city-level ($<$25~km) localization accuracy by 3.6\%--16.58\% and 1.29\%--8.77\%, respectively. Our code and datasets are available : \url{https://github.com/CJ310177/DualGeo}.

\end{abstract}


\section{Introduction}
\label{sec:intro}
Worldwide image geo-localization refers to the task of estimating the geographic coordinates of a photograph captured anywhere on Earth~\cite{1}, and holds significant value across real-world applications. For instance, in digital forensics~\cite{2}, rapidly localizing images shared on social media can aid criminal investigations; in intelligent navigation and augmented reality systems~\cite{3}, it enables location awareness without relying on GPS signals. Furthermore, it underpins large-scale geographic information mining~\cite{4}, environmental change monitoring~\cite{5}, and tourism recommendation systems~\cite{6}. Thus, developing a worldwide image geo-localization method that operates robustly across arbitrary locations and conditions carries substantial scientific and societal significance.
\begin{figure}[t]
	\centerline{\includegraphics[width=0.95\linewidth]{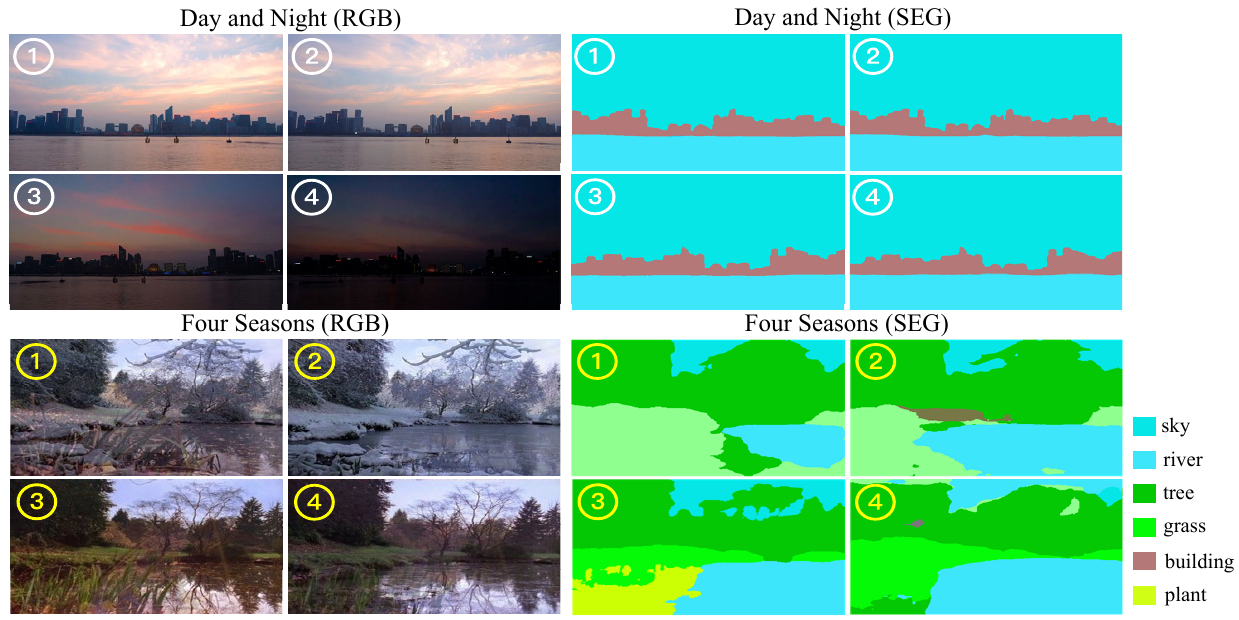}}
	\caption{Scenes under diurnal and seasonal variations. RGB images exhibit significant visual changes, while semantic segmentation maps remain stable.}
	\label{fig1}
	\vspace{-10pt}
\end{figure}

Traditional approaches primarily rely on visual cues in RGB images, such as landmark buildings or natural landscapes, which can effectively support localization under favorable illumination. However, in real-world scenarios, variations in weather, season, and lighting conditions substantially alter image appearance. As illustrated in Fig.~\ref{fig1}, the same location may exhibit drastically different visual characteristics between day and night or across seasons. In contrast to raw visual appearance, the structural layout represented by semantic segmentation maps is more stable and effectively mitigates disruptive changes caused by appearance variations. Existing methods model localization based solely on RGB features; due to the vast diversity and environmental complexity of worldwide imagery, some restrict their scope to specific regions~\cite{7,8,21}. Although recent deep learning approaches trained on large-scale datasets~\cite{9,10,11} have advanced the task toward a global scale, they still struggle to achieve robust and generalizable localization.This limitation primarily manifests in two aspects:

\noindent \textbf{ISSUE 1: Drastic appearance variation of the same location under diverse environmental conditions.} The visual appearance of a location can change dramatically over time, weather, and lighting, while its semantic structure remains stable. We argue that robust worldwide image geo-localization must be grounded in such invariance. Existing methods exhibit significant limitations when confronted with substantial appearance shifts. This inconsistency directly undermines localization strategies relying solely on visual feature matching, leading to markedly reduced robustness in complex environments and frequent erroneous matches.

\noindent \textbf{ISSUE 2: Lack of geo-cognitive post-processing for retrieval results.} Current mainstream approaches typically adopt a nearest-neighbor retrieval paradigm. Even with advanced feature representations learned via deep networks, they often resort to simplistic strategies, such as selecting the \textit{top-1} match or applying basic voting, without leveraging geographic priors. In reality, correct matches tend to form spatial clusters due to the local coherence of geographic space, whereas erroneous candidates appear as outliers. However, existing methods generally lack a geo-cognitive refinement mechanism to analyze the spatial distribution of retrieved GPS candidates, failing to identify and suppress outliers based on clustering patterns. Consequently, their localization results struggle to achieve global-scale generalization and stability.

To address the aforementioned issues, we propose DualGeo, a two-stage framework for worldwide image geo-localization that leverages the invariance of semantic structure and integrates geographic cognition with reasoning. 

\textbf{For ISSUE 1}, DualGeo constructs a geo-representational foundation in the first stage by fusing features from RGB images and their semantic segmentation maps via bidirectional cross-attention, and aligning each view with GPS coordinates through dual-view contrastive learning. This yields a retrieval database enriched with structural semantics. To enable dual-view training, we further generate corresponding semantic segmentation maps for the original MP16 dataset~\cite{18}, resulting in a new variant termed MP16-SEG. \textbf{For ISSUE 2}, DualGeo performs geo-cognitive refinement in the second stage by introducing geographic prior–driven clustering for candidate re-ranking and leveraging large multimodal models (LMMs) for contextual inference to produce the final location estimate. 

The main contributions of this work include:

\begin{itemize}
	\item We propose DualGeo, a two-stage framework for worldwide image geo-localization. It addresses appearance sensitivity by leveraging the structural invariance of semantic segmentation, and boosts localization reliability via geographic clustering and LMMs. DualGeo consistently outperforms existing methods, including GeoCLIP~\cite{14}, PEGION~\cite{13}, and G3~\cite{16}, on three benchmark datasets.
	
	\item We construct MP16-SEG, a large-scale dataset of 4.12M semantic segmentation maps aligned with MP16, annotated with stable structures to provide consistent semantic cues for geo-localization.
	
	\item We design a bidirectional cross-attention mechanism to align RGB images and semantic segmentation maps, improving robustness to appearance changes and yielding more discriminative representations for retrieval.
\end{itemize}

\section{Related work}
\label{sec:Related work}
\subsection{Worldwide Image Geo-localization}
The objective of worldwide image geo-localization is to predict the geographic coordinates of an image’s capture location using a single input image. Existing approaches can be broadly categorized into three classes: classification-based methods, retrieval-based methods, and retrieval-augmented generation (RAG)-based methods. (1) Classification-based methods \cite{9,10,11,12,13} partition the Earth’s surface into fixed spatial units, framing the localization task as a multi-class classification problem. While these methods enable efficient training on large-scale datasets, their accuracy is fundamentally constrained by the spatial granularity of the predefined units: even if the model correctly identifies the target unit, substantial localization errors arise when the true location is near the unit boundary or distant from its centroid. (2) Retrieval-based methods \cite{1,14} perform comparisons between the query image and a geotagged reference database, then designate the location of the most similar database sample as the prediction. These methods depend heavily on the database’s spatial coverage and the discriminative power of the extracted features, but they typically adopt the \textit{top-1} retrieval result directly without modeling the spatial correlations among candidate locations. Consequently, localization failures frequently stem from mismatches when confronted with appearance perturbations or spatially sparse database regions. (3) RAG-based methods \cite{15,16} seek to incorporate LMMs for post-processing: they first retrieve a set of candidate locations, then feed these candidates as prompt information into LMMs to generate the final coordinates. Nevertheless, existing RAG strategies usually construct prompts using a single or unfiltered candidate set. This approach neglects the structural priors of geographic space and thus limits the reliability of the inference process.

\subsection{Contrastive Learning}
The core principle of contrastive learning is to learn discriminative representations by minimizing the distance between semantically relevant samples and maximizing the distance between irrelevant counterparts. In recent years, the CLIP model has exhibited powerful cross-modal representation capabilities by aligning image and text features across large-scale image-text pairs. Inspired by this work, Cepeda et al.~\cite{14} pioneered the replacement of CLIP’s text encoder with a GPS coordinate encoder, establishing a shared embedding space that achieves image-location alignment and thus introducing a novel paradigm for worldwide image geo-localization. Since then, numerous studies~\cite{15,16} have optimized feature fusion, location encoding, and training strategies based on this framework, thereby further boosting the performance of worldwide geo-localization systems.

\section{Method}
\label{sec:Method}
\begin{figure*}[t]
	\centerline{\includegraphics[width=0.94\linewidth,trim=0bp 0bp 0bp 0bp,clip=ture]{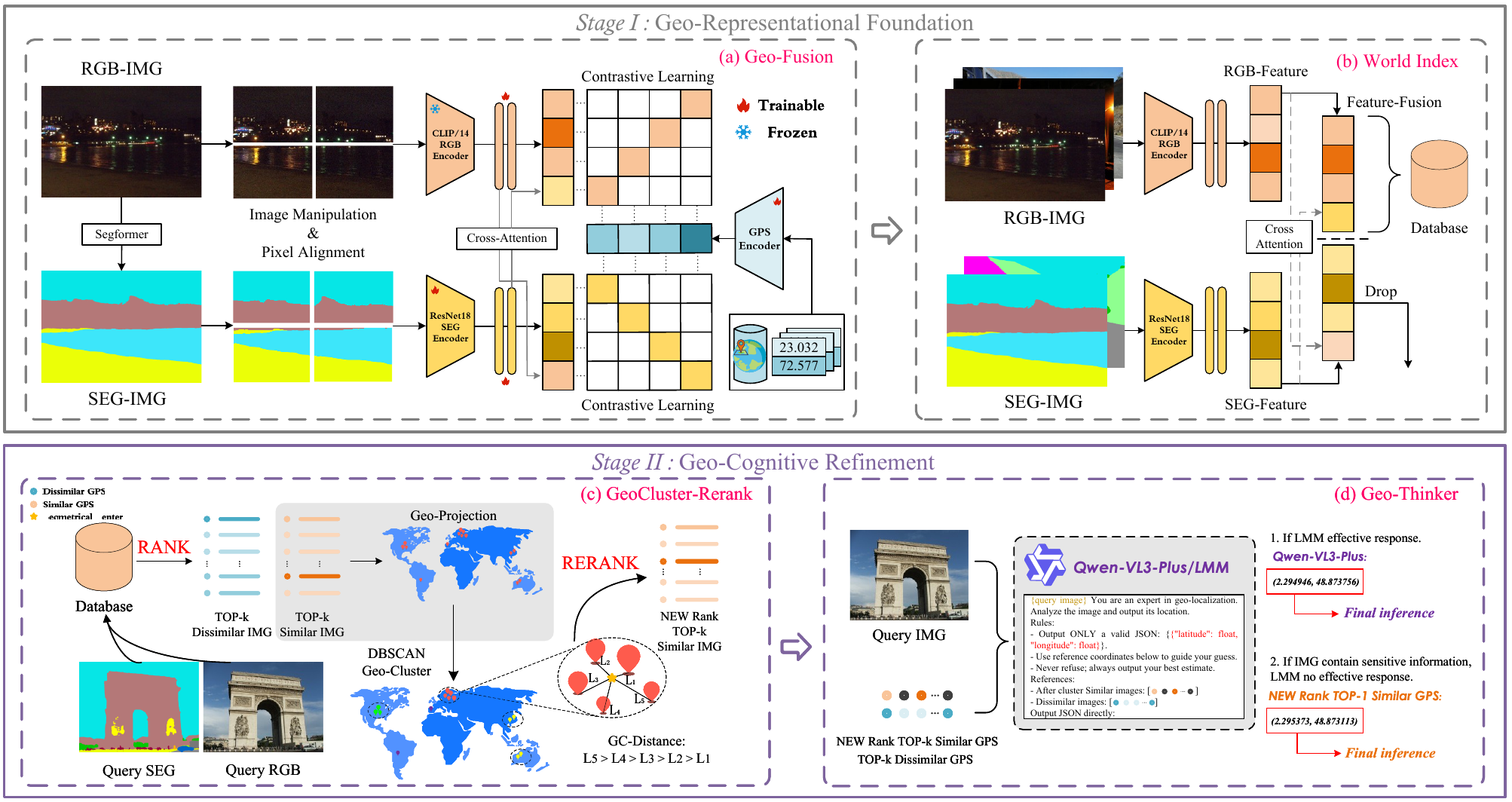}}
	\vspace{-5pt}
	\caption{Overview of the proposed framework DualGeo.}
	\label{fig2}
	\vspace{-5pt}
\end{figure*}

Fig.~\ref{fig2} illustrates the overall framework of DualGeo, which consists of two stages: the \textbf{\textit{geo-representational foundation}} and the \textbf{\textit{geo-cognitive refinement}}. The \textit{geo-representational foundation} integrates two steps: (a)~\textbf{Geo-Fusion}, a dual-view contrastive learning framework that aligns RGB images and their corresponding semantic segmentation maps with GPS coordinates; and (b)~\textbf{World Index}, a global retrieval database built from the fused multimodal features. The subsequent \textit{geo-cognitive refinement} stage further improves localization accuracy through: (c)~\textbf{GeoCluster-Rerank}, which performs geographic clustering of initial retrieval results followed by cluster-aware re-ranking; and (d)~\textbf{Geo-Thinker}, LMMs-based module that leverages the re-ranked candidates and the original query image to generate the final geo-localization prediction.

\subsection{Dual-view Comparative Learning}
\label{3.1}
In the first stage, DualGeo separately encodes RGB images and semantic segmentation maps via dual-view contrastive learning, and achieves joint embedding with geographic locations using a multi-task contrastive loss function. This process ultimately realizes multimodal alignment for both RGB-to-GPS and SEG-to-GPS tasks, as illustrated in Fig.~\ref{fig2}(a). 

Unlike existing methods that align only RGB images with locations~\cite{14} or text~\cite{16}, our approach explicitly unifies appearance and structural views in a shared geographic semantic space. This enables the model to learn rich visual details from RGB data. It also captures structure-invariant cues from semantic segmentation maps that are robust to environmental variations such as lighting, seasonal shifts, and weather conditions. Through cross-modal collaborative training, DualGeo enhances unimodal representations and generates augmented features for retrieval, establishing a robust foundation for global-scale localization. The RGB and GPS encoders follow the architecture of Jia et al.~\cite{16}.

\textbf{SEG Encoder.} We design a dedicated SEG encoder to process single-channel semantic segmentation maps. It is built based on the ResNet-18 architecture \cite{17}. The first-layer convolution kernel is adjusted to adapt to single-channel input, and the fully connected layer is removed to retain spatial structure information. The output features are projected to align with the dimension of the RGB encoder, ensuring compatibility for cross-modal contrastive learning and fusion. By introducing the bidirectional cross-attention mechanism (Section~\ref{3.2}), dynamic information complementarity between RGB images and semantic views is achieved, promoting contextual fusion and providing effective support for cross-view alignment.

\textbf{Contrastive Learning Loss.} In the first stage, the core goal of Geo-Fusion is to establish consistent associations between RGB images and their corresponding semantic segmentation maps and the geographic coordinates of the shooting location respectively. To this end, we adopt a contrastive learning strategy, encouraging each image (whether from RGB or semantic view) to be closer to its true geographic location in its embedding space, while being far away from other locations. We define the loss functions for the two different modalities as follows:
\begin{equation}
	L_{a,b} = -\frac{1}{N} \sum_{i=1}^{n} \log \left( \frac{\exp(S_{i, y_i})}{\sum_{j=1}^{n} \exp(S_{i,j})} \right),
\end{equation}
where $L_{a,b}$ denotes the unidirectional cross-entropy loss from modality $a$ to modality $b$, $S_{i,j}$ denotes the similarity between $i$-th sample image feature and the $j$-th geographic location feature, and $y_i$ indicates that the positive location for the $i$-th sample is the $i$-th location.

During geo-representation learning, DualGeo enforces bidirectional alignment between both RGB and GPS, and SEG and GPS. The overall loss is therefore defined as: $L_t = \frac{1}{2} \left( L_{\mathrm{RGB},\mathrm{GPS}} + L_{\mathrm{GPS},\mathrm{RGB}} + L_{\mathrm{SEG},\mathrm{GPS}} + L_{\mathrm{GPS},\mathrm{SEG}} \right)$. This symmetric contrastive formulation enables the model to jointly utilize fine-grained visual details and structural semantics, learning geo-representations that are robust to environmental changes.

\subsection{Bidirectional Cross-attention Mechanism}
\label{3.2}
We construct a bidirectional cross-attention mechanism to achieve complementary information fusion between RGB images and semantic segmentation maps. This mechanism generates modality-enhanced features through dual-view feature interaction, providing support for Geo-Fusion.
\begin{figure}[t]
	\centerline{\includegraphics[width=0.9\linewidth]{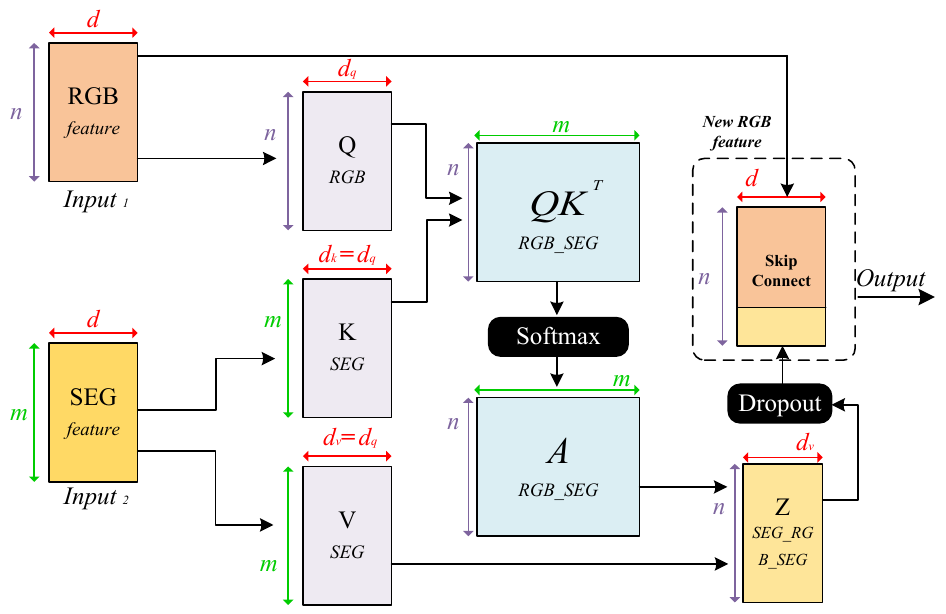}}
	\caption{Schematic diagram of cross-attention. RGB branches.}
	\vspace{-3pt}
	\label{fig3}
	\vspace{-8pt}
\end{figure}

As shown in Fig.\ref{fig3}, the RGB branch incorporates structural cues from semantic segmentation maps to compensate for detail loss in nighttime or low-illumination scenarios, while the semantic branch enhances edge positioning precision by leveraging RGB texture information. For the unidirectional RGB branch, the cross-attention mechanism is formally defined as follows: the query vector is $Q_{\mathrm{RGB}} = W_Q \cdot F_{\mathrm{RGB}}$, the key vector is $K_{\mathrm{SEG}} = W_K \cdot F_{\mathrm{SEG}}$, and the value vector is $V_{\mathrm{SEG}} = W_V \cdot F_{\mathrm{SEG}}$. The attention weight is calculated as:
\begin{equation}
	A_{\mathrm{RGB},\mathrm{SEG}} = \operatorname{Softmax}\left( \frac{Q_{\mathrm{RGB}} \cdot K_{\mathrm{SEG}}^\top}{\sqrt{d_k}} \right),
\end{equation}
with the corresponding output being $Z_{\mathrm{RGB},\mathrm{SEG}} = A_{\mathrm{RGB},\mathrm{SEG}} \cdot V_{\mathrm{SEG}}$. The final enhanced feature is formulated as:
\begin{equation}
	F_{\mathrm{RGB}}^{\mathrm{new}} = \operatorname{LayerNorm}\bigl( \operatorname{Dropout}(Z_{\mathrm{RGB},\mathrm{SEG}}) + F_{\mathrm{RGB}} \bigr),
\end{equation}
where $W_Q, W_K, W_V$ denote learnable projection matrices, and $d_k$ is the shared dimension of query and key vectors. Cross-attention for the SEG branch is obtained by swapping RGB and SEG features, yielding the enhanced representation $F_{\mathrm{SEG}}^{\mathrm{new}}$. 

After bidirectional interaction, the augmented features retain the discriminative strength of each modality and improve cross-view alignment, enabling accurate visual–geographic matching in Geo-Fusion.

\textbf{World Index.} As shown in Fig.\ref{fig2}(b), upon completion of the Geo-Fusion stage, we construct the global index using the enhanced RGB features, where $F' = F_{\mathrm{RGB}}^{\mathrm{new}}$. Here, $F'$ denotes the unique representation of each image in the index, ensuring efficient storage and fast retrieval of the image features.

\subsection{Geo-cognitive Refinement}
\label{3.3}
This section presents the second stage of DualGeo: \textit{geo-cognitive refinement}, which comprises two components. The first is \textbf{GeoCluster-Rerank}, designed for geographically clustered re-ranking (Fig.~\ref{fig2}(c)). The second is \textbf{Geo-Thinker}, which performs LMMs-based reasoning (Fig.~\ref{fig2}(d)).

Using the retrieval database built in the previous stage, we first retrieve the \textit{top-$k$} most visually similar GPS coordinates $S = \{s_j = (\text{lat}_j, \text{lon}_j)\}_{j=1}^k$ for a query image (encoded via RGB and segmentation features). Due to illumination, season, or viewpoint variations, $S$ may contain geographic outliers. Notably, true locations typically form dense local clusters, while mismatches appear as isolated points.

\textbf{GeoCluster-Rerank} proceeds in three steps:

\noindent\textbf{1. Initial Retrieval.} 
First, we obtain $S$ via nearest-neighbor search, along with the \textit{top-$k$} most dissimilar coordinates $D = \{d_j = (\text{lat}_j, \text{lon}_j)\}_{j=1}^k$. Candidates in $S$ are ranked by visual similarity, ignoring geospatial distribution.

\noindent\textbf{2. Geospatial Clustering.} 
Then, we project them onto the world map and apply the DBSCAN algorithm for cluster analysis. Specifically, the coordinates are first converted to radians, the distance metric of DBSCAN adopts the spherical Haversine distance, and its neighborhood radius is denoted as $\varepsilon $, is normalized to spherical radians. DBSCAN can identify high-density regions (i.e., clusters where real shooting locations may exist), and regard coordinates in isolated or low-density regions as outliers. Then, the candidate coordinates $S$ are labeled to obtain the cluster label ${l_j} \in \{  - 1,0,1,...\} $ of each candidate, where ${l_j} = -1$ indicates noise points. The set of all non-noise clusters is denoted as $L = \{ l|l \ne  - 1\} $. For each $l \in L$, its Geometrical Center (GC) is calculated. GC serves as the spatial average of all points in the cluster, providing the best estimate of the potential real location.

\noindent\textbf{3. Re-Ranking.} 
If $L \neq \emptyset$, we select the largest cluster as the main cluster and use its geometric center $\bar{s}_l$ as the reference position; if $L = \emptyset$, we take the mean of all coordinates as $\bar{s} = \frac{1}{k}\sum_{j=1}^k s_j$. Subsequently, we compute the Haversine distance between each candidate position in $S$ and the reference position:
\begin{equation}
	m_j = \mathrm{Haversine}(s_j, \bar{s}_{\text{ref}}),
\end{equation}
where $\bar{s}_{\text{ref}} \in \{\bar{s}_l, \bar{s}\}$ is the reference position. Finally, we re-rank the candidates in ascending order of distance:
\begin{equation}
	\pi = \mathrm{argsort}([m_1, m_2, \dots, m_k]).
\end{equation}

This yields the re-ranked set $S' = \{s_{\pi(1)}, s_{\pi(2)}, \dots, s_{\pi(k)}\}$, where the first element $s_{\pi(1)}$ serves as the \textit{top-1} localization prediction after re-ranking. By shifting the ranking criterion from original feature similarity to geospatial consistency, this strategy effectively suppresses the impact of outlier matches, significantly improving the robustness and rationality of localization without introducing additional models.

To further enhance localization reliability, DualGeo’s \textbf{Geo-Thinker} interprets the retrieval results through geospatial reasoning with LMMs. As shown in Fig.~\ref{fig2}(d), it constructs a contrastive geospatial context by selecting the \textit{top-$n$} most similar coordinates from the re-ranked set $S'$ and the \textit{top-$n$} most dissimilar coordinates from the initial dissimilar set $D$ (where $n < k$). A structured prompt, comprising the query image and these $2n$ coordinates, is fed to LMMs. The model fuses visual and world knowledge to filter out unreasonable candidates, acting as a geospatial arbiter rather than a regressor, thereby improving both robustness and interpretability in global geo-localization~\cite{20}. The prompt construction for LMMs is detailed in the supplementary material.

\section{Experiments}

\subsection{Datasets and Experimental Setup}
\textbf{MP16-SEG.} To enhance DualGeo's robustness to lighting, seasonal, and weather variations, we generated high-resolution semantic segmentation maps for the MP16 dataset~\cite{18}, creating MP16-SEG. Using a SegFormer-B5 model~\cite{19} pretrained on the ADE20K, we segmented images into 150 semantic categories (e.g., sky, buildings, vegetation). Semantic segmentation layouts remain stable across environmental changes, making them more suitable than RGB appearances as geographically invariant features. MP16-SEG provides structured semantic priors, enabling the model to reduce reliance on RGB appearance during localization and achieve more robust and flexible modeling through semantic consistency.

\textbf{Datasets.} We utilize the MP16 \cite{18} and MP16-SEG datasets to build the training set and retrieval library. The test sets consist of three components: the IM2GPS dataset \cite{1}, a manually curated few-shot dataset; and two standard benchmarks for worldwide image geo-localization, namely the IM2GPS3k and YFCC4k datasets \cite{4}. Corresponding semantic segmentation maps are generated for all test sets.

\textbf{Evaluation Metrics.} We adopt the spherical distance metric, calculating the proportion of samples where the distance between the predicted location and ground-truth coordinates falls within the following thresholds: 1 km (street-level), 25 km (city-level), 200 km (region-level), 750 km (country-level), and 2500 km (continent-level).

\textbf{Experimental Details.} All experiments are implemented based on PyTorch. RGB and semantic segmentation features are unified to a dimensionality of 768. Training is conducted for 10 epochs using the AdamW optimizer, with a batch size of 256, an initial learning rate of $4\mathrm{e}{-5}$, and dynamic weight decay. The temperature coefficient for contrastive learning is uniformly set to 3.99. For \textit{top-k} nearest neighbor search, we set $k=20$. In geospatial clustering, the neighborhood radius is configured as $\varepsilon = 5\ \text{km}$. LMM is performed using the \textit{Qwen3-VL-Plus} model, with $n=10$ for IM2GPS, $n=15$ for IM2GPS3k, and $n=5$ for YFCC4k. More training details are provided in the supplementary material.

\subsection{Comparison with State-of-the-art Methods}
To assess the geo-localization performance of DualGeo, we conduct comprehensive experiments on the three aforementioned test sets. The baseline methods selected for comparison include: kNN $\sigma=4$ \cite{4}, PlaNet \cite{9}, CPlaNet \cite{10}, ISN \cite{11}, Translocator \cite{12}, GeoCLIP \cite{14}, Img2Loc \cite{15}, PIGEON \cite{13}, and G3 \cite{16}. The detailed descriptions of baselines are in supplementary material. Experimental results for different methods are presented in Tab.~\ref{tab:results}. DualGeo achieves strong fine-grained localization performance, especially at the street and city levels, due to the effective fusion of semantic structure and geographic reasoning, which enhances robustness to environmental appearance variations. A slight drop in coarse-scale accuracy (e.g., country or continent) is observed, as the model emphasizes local structural consistency and prioritizes precise localization over global ambiguity resolution. Overall, DualGeo delivers superior performance across multiple datasets and demonstrates remarkable fine-grained inference capability.
\begin{table}[t]
	\caption{Main results on IM2GPS, IM2GPS3k and YFCC4k.}
	\vspace{-5pt}
	\label{tab:results}
	\centering
	\adjustbox{max width=0.95\linewidth}{
		\begin{tabular}{l l c c c c c}
			\toprule
			\multirow{2}{*}{Benchmark} & \multirow{2}{*}{Method} & \multicolumn{5}{c}{Distance (\% @ km)} \\
			\cmidrule(lr){3-7}
			& & \makecell{Street\\1 km} & \makecell{City\\25 km} & \makecell{Region\\200 km} & \makecell{Country\\750 km} & \makecell{Continent\\2,500 km} \\
			\midrule
			\multirow{6}{*}{IM2GPS \cite{1}} 
			& PlaNet \cite{9}& 8.4 & 24.5 & 37.6 & 53.6 & 71.3 \\
			& CPlaNet \cite{10}& 16.5 & 37.1 & 46.4 & 62.0 & 78.3 \\
			& ISNs \cite{11}& 16.9 & 43.0 & 51.9 & 66.7 & 80.2 \\
			& Translocator \cite{12}& \underline{19.8} & \underline{48.1} & \underline{64.6} & 75.6 & 86.7 \\
			& PIGEON \cite{13}& 14.8 & 40.9 & 63.3 & \underline{82.3} & \underline{91.1} \\
			\cmidrule{2-7}
			& DualGeo(ours) & \textbf{\itshape 23.2} & \textbf{\itshape 52.32} & \textbf{\itshape 65.4} & 78.9 & 89.87 \\
			& \textit{Improvement (\%)} & +16.58 & +8.77 & +1.23 & -4.13 & -1.35 \\
			\midrule
			\multirow{6}{*}{IM2GPS3k \cite{4}} 
			& kNN,$\sigma$=4 \cite{4} & 7.2 & 19.4 & 26.9 & 38.9 & 55.9 \\
			& PlaNet \cite{9} & 8.5 & 24.8 & 34.3 & 48.4 & 64.6 \\
			& CPlaNet \cite{10}& 10.2 & 26.5 & 34.6 & 48.6 & 64.6 \\
			& ISNs \cite{11}& 10.5 & 28.0 & 36.6 & 49.7 & 66.0 \\
			& Translocator \cite{12}& 11.8 & 31.1 & 46.7 & 58.9 & 76.1 \\
			& GeoCLIP \cite{14}& 14.11 & 34.47 & 50.65 & 69.67 & 83.82 \\
			& Img2Loc \cite{15}& 15.34 & 39.83 & 53.39 & 69.7 & 82.78 \\
			& PIGEON \cite{13}& 11.3 & 36.7 & 53.8 & \underline{72.4} & \underline{85.3} \\
			& G3 \cite{16}& \underline{16.65} & \underline{40.94} & \underline{55.56} & 71.24 & 84.68 \\
			\cmidrule{2-7}
			& DualGeo(ours) & \textbf{\itshape 17.25} & \textbf{\itshape 41.47} & \textbf{\itshape 55.76} & 71.71 & 85.05 \\
			& \textit{Improvement (\%)} & +3.60 & +1.29 & +0.36 & -0.95 & -0.29 \\
			\midrule
			\multirow{6}{*}{YFCC4k~\cite{4}} 
			& kNN,$\sigma$=4 \cite{1}& 2.3 & 5.7 & 11.0 & 23.5 & 42.0 \\
			& PlaNet \cite{9}& 5.6 & 14.3 & 22.2 & 36.4 & 55.8 \\
			& CPlaNet \cite{10}& 7.9 & 14.8 & 21.9 & 36.4 & 55.5 \\
			& ISNs \cite{11}& 6.5 & 16.2 & 23.8 & 37.4 & 55.0 \\
			& Translocator \cite{12}& 8.4 & 18.6 & 27.0 & 41.1 & 60.4 \\
			& GeoCLIP \cite{14}& 9.59 & 19.31 & 32.63 & 55.0 & 74.69 \\
			& Img2Loc \cite{15}& 19.78 & 30.71 & 41.4 & 58.11 & 74.07 \\
			& PIGEON \cite{13}& 10.4 & 23.7 & 40.6 & 62.2 & 77.7 \\
			& G3 \cite{16}& \underline{23.99} & \underline{35.89} & \underline{46.98} & \underline{64.26} & \underline{78.15} \\
			\cmidrule{2-7}
			& DualGeo(ours) &\textbf{27.49}&\textbf{36.45} &45.03& 61.58 & 75.92\\
			& \textit{Improvement (\%)} & +14.59 & +1.56 & -4.15 & -4.17 & -2.85 \\
			\bottomrule
		\end{tabular}
	}
	\vspace{-8pt}
\end{table}

\subsection{Ablation Study}
\begin{table}[b]
	\vspace{-8pt}
	\centering
	\caption{Ablation study on IM2GPS3k.}
	\vspace{-5pt}
	\label{tab:ablation}
	\setlength{\tabcolsep}{4pt} 
	\adjustbox{max width=0.95\linewidth}{
	\begin{tabular}{l c c c c c}
		\toprule
		Method
		& \makecell{Street\\1 km} & \makecell{City\\25 km} & \makecell{Region\\200 km} & \makecell{Country\\750 km} & \makecell{Continent\\2,500 km} \\
		\midrule
		$M_{RGB}$ & 11.11 & 29.83 & 40.61 & 55.86 & 73.41 \\
		$M_{Dual}$ & 12.25 & 31.97 & 42.64 & 57.59 & 75.19 \\
		$M_{Dual\_CT}$ & 12.81 & 32.3 & 43.24 & 58.79 & 76.37 \\
		$M_{Dual\_CT\_RR5}$ & 14.81 & 34.5 & 45.41 & 60.19 & 76.78 \\
		$M_{Zero\_LMM}$ & 8.54 & 33.13 & 52.12 & 70.84 & 83.32 \\
		\cmidrule{1-6}
		Full model & \textbf{17.25} & \textbf{41.47} & \textbf{55.76} & \textbf{71.71} & \textbf{85.05} \\
		\bottomrule
	\end{tabular}}
\end{table}

To verify the effectiveness of each component of DualGeo, we conduct ablation experiments on the IM2GPS3k dataset. We design the following variant models: $M_{\mathrm{RGB}}$: training with only RGB images for contrastive learning; $M_{\mathrm{Dual}}$: training with dual-view contrastive learning instead; $M_{\mathrm{Dual\_CT}}$: adding bidirectional cross-attention mechanism on the basis of $M_{\mathrm{Dual}}$; $M_{\mathrm{Dual\_CT\_RR5}}$: adding the geographic clustering re-ranking mechanism on the basis of $M_{\mathrm{Dual\_CT}}$, setting $\varepsilon = 5\ \text{km}$; $M_{\mathrm{Zero\_LMM}}$: relying only on the LMM for image location inference; Full model: the complete DualGeo model.
\begin{figure*}[t]
	\centerline{\includegraphics[width=0.95\linewidth,trim=10bp 10bp 10bp 10bp,clip=ture]{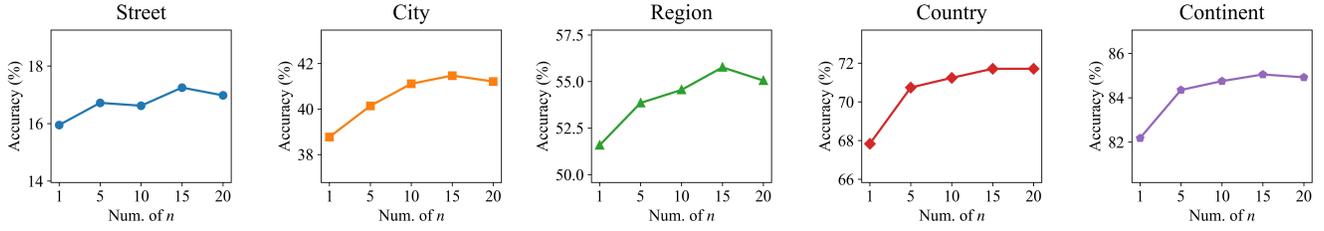}}
	\vspace{-5pt}
	\caption{Performance of different $n$ values on LMM on IM2GPS3k dataset.}
	\label{fig5}
	\vspace{-5pt}
\end{figure*}

In Tab.~\ref{tab:ablation}, we find that (1) every component contributes substantially to DualGeo’s final performance; (2) geographic clustering significantly boosts fine-grained localization accuracy; (3) the LMM’s extensive knowledge base enables strong reasoning at country and continental scales; and (4) the complete DualGeo effectively integrates the fine-grained precision from clustering with the LMM’s global contextual awareness, yielding balanced performance across all scales.

We perform LMMs hyperparameter analysis on $n \in \{1,5,10,15,20\}$, shown in Fig.~\ref{fig5}. Performance quickly saturates as $n$ increases, owing to geospatial prior-guided re-ranking, which refines candidate ordering, filters implausible coordinates, and conserves token budget. Ablations with different SEG encoders are provided in the supplementary material.

\begin{figure}[t]
	\centerline{\includegraphics[width=0.93\linewidth,trim=10bp 10bp 10bp 10bp,clip=ture]{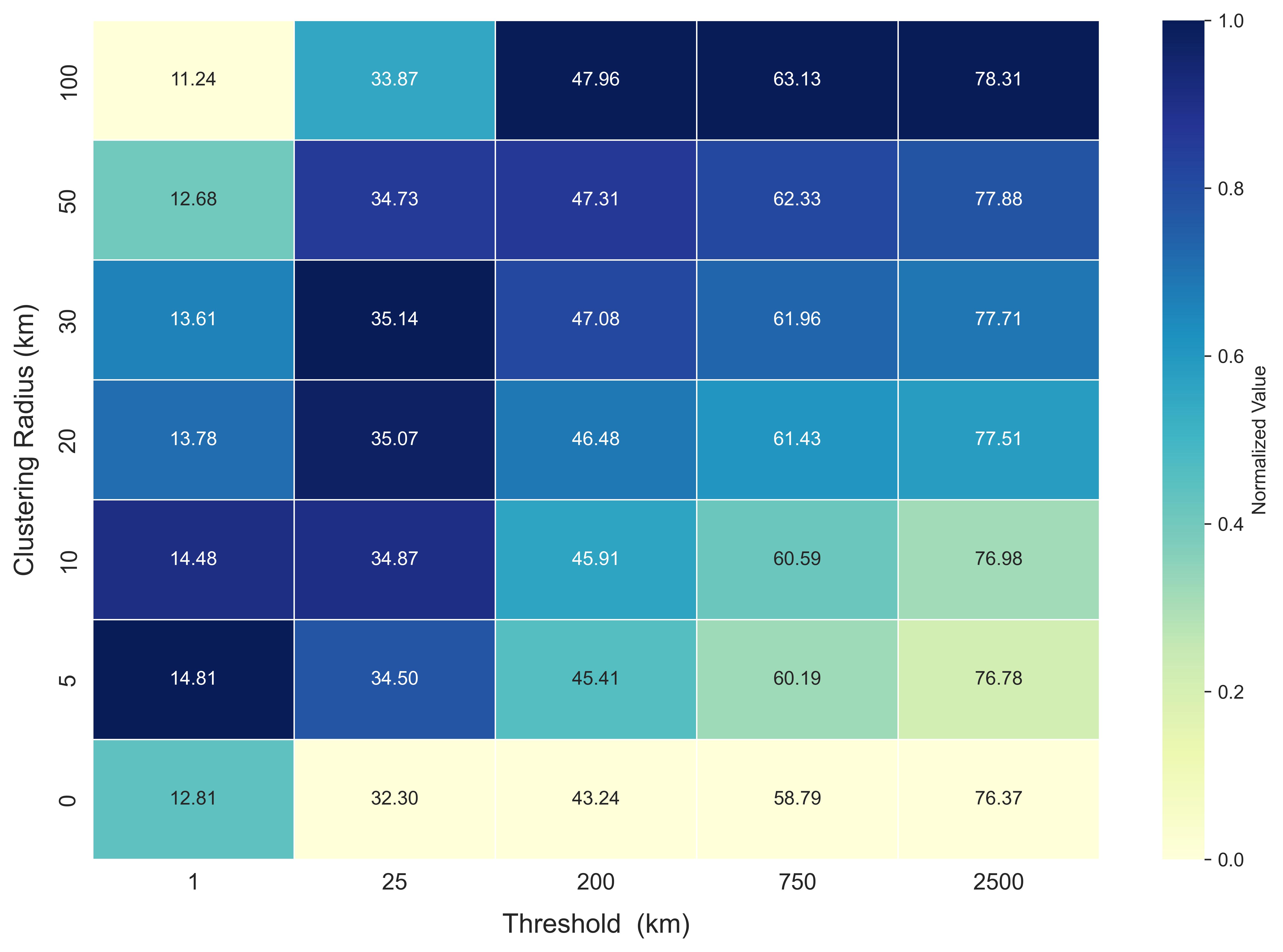}}
	\caption{Performance of different clustering radii on the IM2GPS3k dataset.}
	\label{fig4}
	\vspace{-4pt}
\end{figure}

We also conduct an ablation study on the impact of re-ranking clustering radius $\varepsilon \in \{0,5,10,20,30,50,100\}\ \text{km}$ on geo-localization performance, shown in Fig.~\ref{fig4}, with $M_{\mathrm{Dual\_CT}}$ ($\varepsilon=0\,\text{km}$) as baseline. Results show that a smaller clustering radius yields higher accuracy in fine-grained localization, whereas increasing it effectively improves performance at long distances. The choice of radius depends on target scenario distance requirements.

\section{Conclusion}
Existing worldwide image geo-localization methods rely heavily on visual appearance and are vulnerable to lighting, seasonal, and weather changes. We propose DualGeo, a two-stage framework that first fuses RGB and semantic features via dual-view contrastive learning and bidirectional attention to build a retrieval database, and then re-ranks candidates using geographic clustering before feeding the re-ranked results together with the original image into LMMs for final localization. Experiments show that semantic segmentation mitigates environmental appearance variations, while geographic clustering significantly boosts localization accuracy.

\section{Acknowledgment}

This work was supported in part by the National Natural Science Foundation of China  (No. U23A20305, No. 62372465), Natural Science Foundation of Henan Province (No. 262300422589), and the Innovation Scientists and Technicians Troop Construction Projects of Henan Province (No. 254000510007).

\bibliographystyle{IEEEbib}
\bibliography{icme2026references}

\end{document}